\documentclass[letterpaper]{article} 
\usepackage{aaai20}  
\usepackage{times}  
\usepackage{helvet} 
\usepackage{courier}  
\usepackage[hyphens]{url}  
\usepackage{graphicx} 
\usepackage{caption}
\usepackage{subcaption}
\usepackage{graphicx}  
\usepackage{xcolor}

\newcommand{\hide}[1]{}
\usepackage{bbm}
\usepackage{amsmath,amssymb,amsthm}
\usepackage{multirow}

\newcommand{\blackbox}{B}
\newcommand{\explanation}{E}
\newcommand{\blackboxes}{\mathcal{B}}
\newcommand{\explanations}{\mathcal{E}}
\newcommand{\para}[1]{\vspace{4pt}\noindent\textbf{#1}}

\newcommand{\X}{\mathcal{X}}
\newcommand{\Y}{\mathcal{Y}}
\newcommand{\F}{\mathcal{E}}

\newcommand{\OO}{\mathcal{O}}

\newcommand{\XP}{\mathcal{X}_P}
\newcommand{\XA}{\mathcal{X}_A}
\newcommand{\XD}{\mathcal{X}_D}

\newcommand{\FT}{\tilde{\F}}

\newcommand{\ft}{\tilde f}

\newcommand{\argmax}{\operatorname*{\arg\max}}

\newtheorem{theorem}{Theorem}[section]

\title{``How do I fool you?": Manipulating User Trust \\via Misleading Black Box Explanations}
\author{Himabindu Lakkaraju*, Osbert Bastani$^+$\\
*Harvard University, $^+$University of Pennsylvania\\
\{*hlakkaraju@seas.harvard.edu, $^+$obastani@seas.upenn.edu\}}

\begin{document}

\maketitle

\begin{abstract}
As machine learning black boxes are increasingly being deployed in critical domains such as healthcare and criminal justice, there has been a growing emphasis on developing techniques for explaining these black boxes in a human interpretable manner.
It has recently become apparent that a high-fidelity explanation of a black box ML model may not accurately reflect the biases in the black box.
As a consequence, explanations have the potential to mislead human users into trusting a problematic black box. In this work, we rigorously explore the notion of misleading explanations and how they influence user trust in black box models. More specifically, we propose a novel theoretical framework for understanding and generating misleading explanations, and carry out a user study with domain experts to demonstrate how these explanations can be used to mislead users.
Our work is the first to empirically establish how user trust in black box models can be manipulated via misleading explanations.
\end{abstract}

\section{Introduction}
There has been an increasing interest in using ML models to aid decision makers in domains such as healthcare and criminal justice. In these domains, it is critical that decision makers understand and trust ML models, to ensure that they can diagnose errors and identify model biases correctly. However, ML models that achieve state-of-the-art accuracy are typically complex \emph{black boxes} that are hard to understand. As a consequence,
there has been a recent surge in post hoc explanation techniques for explaining black box models~\cite{ribeiro2018anchors,ribeiro16:kdd,lakkaraju19faithful,lundberg2017unified}. One of the goals of such explanations is to help domain experts detect systematic errors and biases in black box model behavior~\cite{doshi2017towards}.

Existing techniques for explaining black boxes typically rely on optimizing \emph{fidelity}---i.e., ensuring that the explanations accurately mimic the predictions of black box model~\cite{ribeiro16:kdd,ribeiro2018anchors,lakkaraju19faithful}. 
The key assumption underlying these approaches is that if an explanation has high fidelity, then biases of the black box model will be reflected in the explanation. However, it is questionable whether this assumption actually holds in practice~\cite{lipton2016mythos}.
The key issue is that high fidelity \emph{only} ensures high correlation between the predictions of the explanation and the predictions of the black box. There are several other challenges associated with post hoc explanations which are not captured by the fidelity metric: (i) they may fail to capture causal relationships between input features and black box predictions~\cite{lipton2016mythos,rudin2019stop},
(ii) there could be multiple high-fidelity explanations for the same black box that look qualitatively different~\cite{lakkaraju19faithful},
and (iii) they may not be robust and can vary significantly even with small perturbations to input data~\cite{ghorbani2019interpretation}.

These challenges increase the possibility that explanations generated using existing techniques can actually \emph{mislead} the decision maker into trusting a problematic black box. However, there has been little to no prior work empirically studying if and how explanations can mislead users.

\begin{figure*}
\centering
\includegraphics[width=0.40\textwidth]{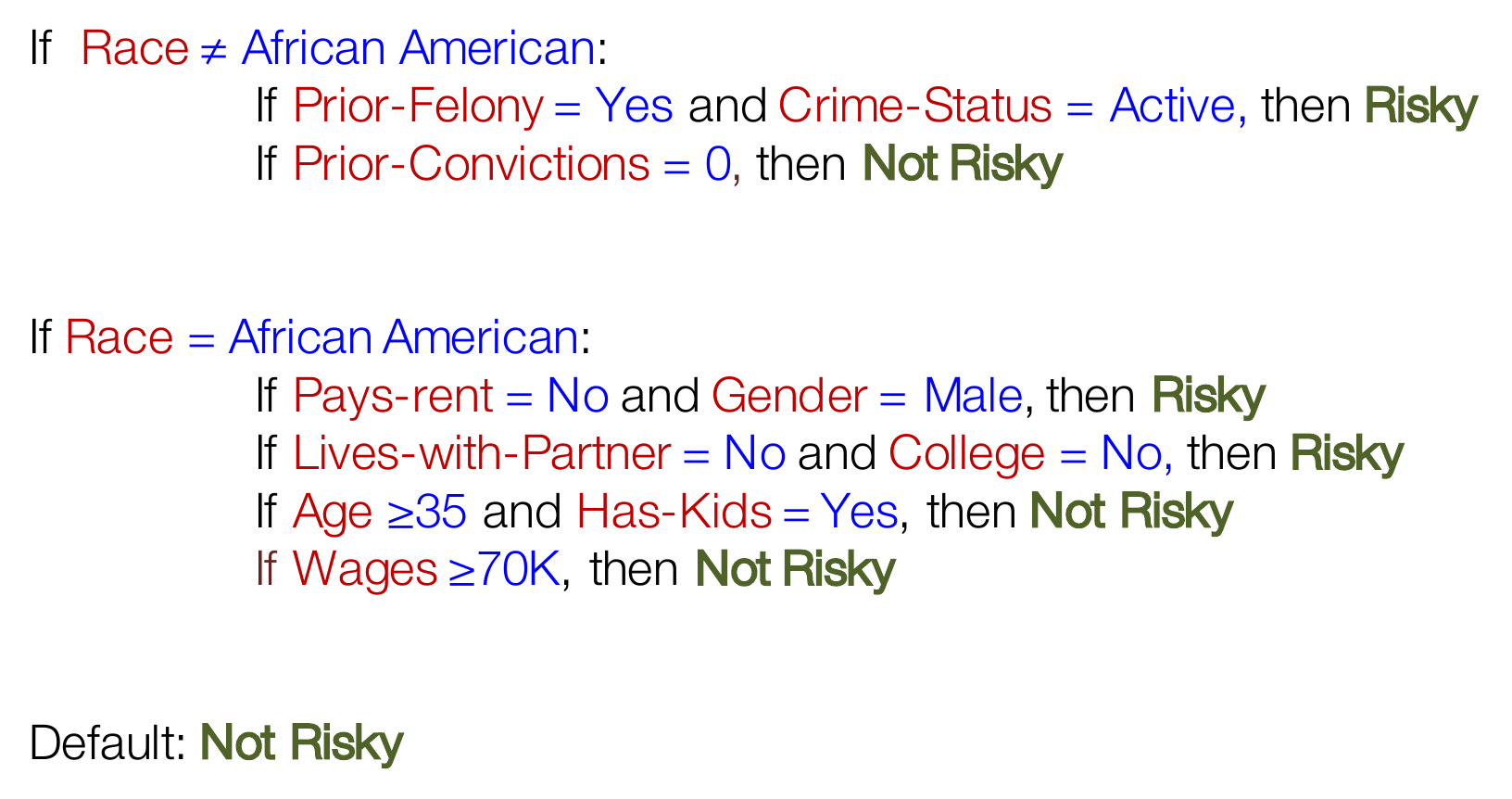}
\hspace{0.3in}
\includegraphics[width=0.50\textwidth]{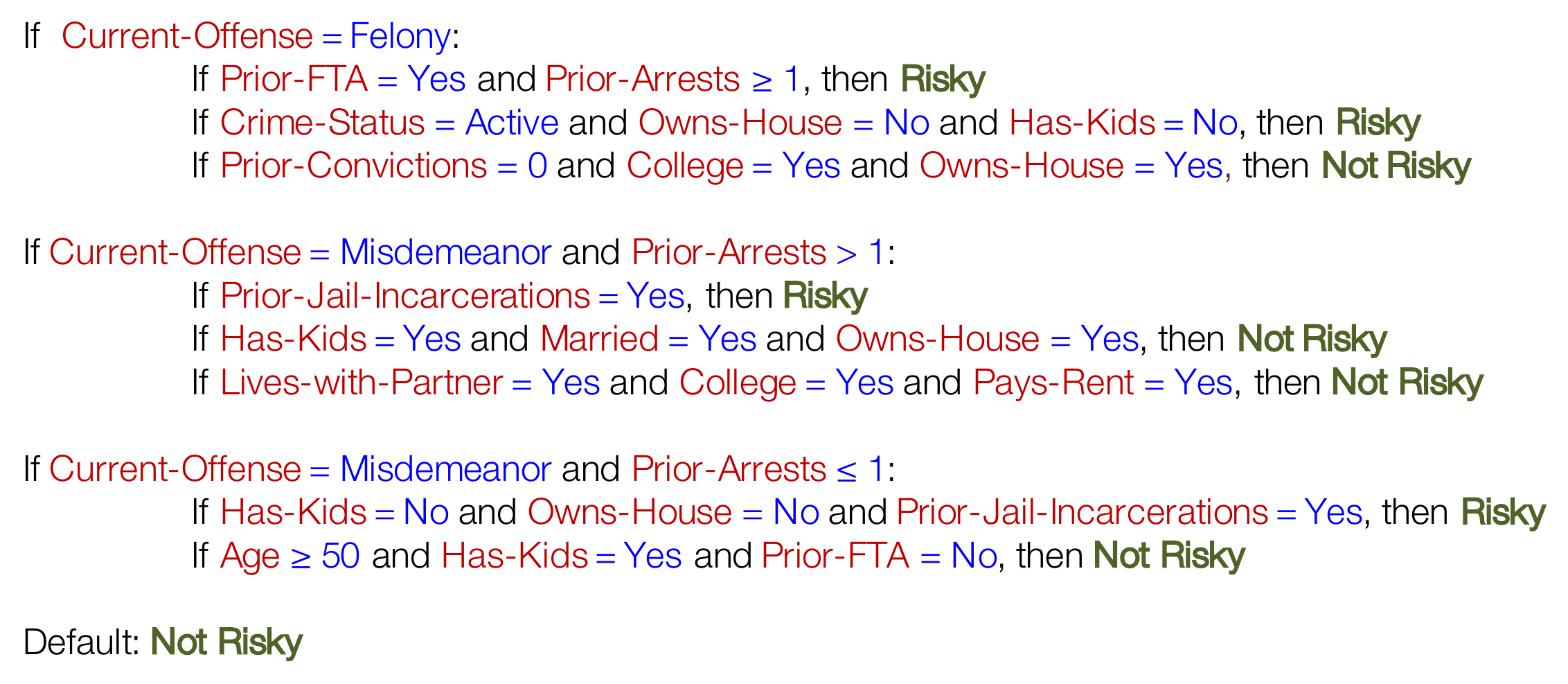}
\vspace{-0.1in}
\caption{ Classifier which uses prohibited features (race and gender) when making predictions (left); and its \emph{misleading} explanation (right) which excludes prohibited features (race, gender) and includes desired features (prior jail incarcerations, prior FTA\footnote{flight risk}). Our user study shows that domain experts are 9.8 times more likely to trust the classifier if they see the explanation on the right (instead of the classifier). Presence or absence of race and gender drives user trust (See Section~\ref{sec:evaluation}) }
\label{fig:blackboxandexp}
\vspace{-0.2in}
\end{figure*}
\para{Contributions.}
We propose the first systematic study to explore if and how explanations of black boxes can mislead users. First, we propose a novel theoretical framework for understanding when misleading explanations can exist. We show that even if an explanation achieves perfect fidelity, it may still not reflect issues in the black box model.
The key issue is that due to correlations in the features, explanations can achieve high fidelity even if they use entirely different features compared to the black box.
Second, we propose a novel approach for generating \emph{potentially} misleading explanations. Our approach extends the MUSE framework~\cite{lakkaraju19faithful} to favor explanations that contain features that users believe are relevant and omit features that users believe are problematic.
Third, we perform an extensive user study with domain experts from law and criminal justice to understand how misleading explanations impact user trust. Our results demonstrate that the misleading explanations generated using our approach
can in fact increase user trust of by 9.8 times (See Figure~\ref{fig:blackboxandexp}).
Our findings have far reaching implications both for research on ML interpretability and real-world applications of ML.

\para{Related work.}
%
Present work on interpretable ML largely falls into three categories. First, there are approaches focused on learning predictive models that are human understandable~\cite{letham15:interpretable,lakkaraju16:interpretable,caruana15:intelligible}.
However, complex models such as deep neural networks and random forests typically achieve higher performance compared to interpretable models~\cite{ribeiro16:kdd}, so in many situations it is more desirable to use these complex models. Thus, there has been work on explaining such complex black boxes.
One approach is to provide local explanations for individual predictions of the black box~\cite{ribeiro2018anchors,ribeiro16:kdd,lundberg2017unified}, which is useful when a decision maker plans to review every decision made by the black box. An alternate approach is to provide a global explanation that describes the black box as a whole, typically summarizing it using an interpretable model~\cite{lakkaraju19faithful,bastani2017interpretability}, which is useful in validating the black boxes before they are deployed to automatically make decisions (i.e., without human involvement).

There has been some empirical work on studying how humans understand and trust interpretable models and explanations. For instance, Poursabzi-Sangdeh et. al. (2018) show that longer explanations are harder for humans to simulate accurately.
There has also been recent work on understanding what makes explanations useful in the context of three tasks they are likely to perform given an explanation of an ML system: (i) predicting the system's output, (ii) verifying whether the output is consistent with the explanation, and (iii) determining if and how the output would change if we change the input~\cite{lage2019evaluation}.

More closely related to our work, there has been recent work on exploring the vulnerabilities of black box explanations. For instance, there has been work demonstrating that explanations can be unstable, changing drastically even with small perturbations to inputs~\cite{dombrowski2019explanations,ghorbani2019interpretation}. Finally, recent work has argued that black box explanations can often be misleading and can potentially lead users to trust problematic black boxes~\cite{lipton2016mythos,ghorbani2019interpretation}.

In contrast, we are the first to study if and how adversarial entities could generate misleading explanations to manipulate user trust. 
We are also the first to explore the notion of confirmation bias in the context of black box explanations.

\hide{user studies about interpretability\\
-- HCOMP 2019, Isaac, Finale\\
-- Porsabzi Sangdeh\\

interpretable models with user studies\\
-- Lakkaraju 2016, Kim 2015 etc. \\

black box explanations with user studies\\
-- LIME\\
--SHAP\\
--Lakkaraju 2019\\

Issues with black box posthoc explanations\\
\cite{ghorbani2019interpretation}

\cite{rudin2019stop}
}

\section{Problem Formulation}

In this section, we introduce some notation and formalize the notions of (i) explanation of a black box model, and (ii) misleading explanation of a black box model. 

\para{Explanations.}
Given input data $\X$, a set of class labels $\Y = \{1, 2, \cdots K\}$, and a black box $\blackbox:\X\to\Y$, our goal is to generate an \textbf{\emph{explanation}} $\explanation$ that describes the behavior of
$\blackbox$. Then, end users can use $\explanation$ to determine whether to trust $\blackbox$.

We consider an approach to explaining $\blackbox$ by approximating it using an interpretable model $\explanation\in\explanations$. We measure the quality of this approximation using the \emph{relative error}
\begin{align*}
L(\explanation,\blackbox)=\mathbb{E}_{p(x)}[\ell(\explanation(x),\blackbox(x))]
\end{align*}
where $p(x)$ is the data distribution and $\ell(y,y')$ is any loss function---e.g., the 0-1 loss $\ell(y,y')= \mathbb{I}[y\neq y']$. We want to choose an explanation $\explanation\in\explanations$ that minimizes the relative error. We also define the \emph{fidelity} of $\explanation$ to be $1-L(\explanation,\blackbox)$.

\para{Trustworthy black boxes \& misleading explanations.}
%
We assume a workflow where the human user relies on $\hat\explanation$ to decide whether to trust $\blackbox$. We model the human user as an oracle $\OO:\explanations\to\{0,1\}$ such that
\begin{align*}
\OO(\explanation)=\mathbb{I}[\text{user trusts black box }\blackbox \text{ given explanation }\explanation].
\end{align*}
We can compute $\OO$ via a user study that shows users $\hat\explanation$ and asks if they trust $\blackbox$. We also assume there is a ``correct'' choice of whether $\blackbox$ is \textbf{\emph{trustworthy}}. We model this ground truth as an oracle $\OO^*:\blackboxes\to\{0,1\}$, where $\blackboxes$ is the space of all black boxes and
$\OO^*(\blackbox)=\mathbb{I}[\blackbox \text{ is trustworthy}]$.
An explanation $\explanation$ for $\blackbox$ is \textbf{\emph{misleading}} if $\OO(\explanation)\neq\OO^*(\blackbox)$.

\para{Constructing misleading explanations.}
Our goal is to demonstrate that misleading explanations exist. In our approach, we first devise a black box $\blackbox$ that we expect to be untrustworthy. This expectation is based on which features are used by the model (see Section~\ref{sec:theory}). Then, we need to check if $\blackbox$ is \emph{actually} untrustworthy (i.e., $\OO^*(\blackbox)=0$). To do so, we choose $\blackbox$ to itself be an interpretable model. Then, we perform a user study where we show $\blackbox$ and ask if it is trustworthy, yielding $\OO^*(\blackbox)$. In this approach, $\blackbox$ is still a black box in the sense that (i) $\hat\explanation$ is constructed without examining the internals of $\blackbox$, and (ii) users are not aware of the internals of $\blackbox$ when shown $\explanation$ to evaluate $\OO(\explanation)$.

Next, we construct an explanation $\explanation$ of $\blackbox$ that we expect to be misleading; again, this expectation is based on which features are in the explanation (see Section~\ref{sec:theory}). Then, we check if $\explanation$ is indeed misleading (i.e., evaluate $\OO(\explanation)$) via a user study. Assuming we successfully constructed $\blackbox$ so that $\OO^*(\blackbox)=0$, then $\explanation$ is misleading if $\OO(\explanation)=1$. We discuss how we construct $\explanation$ in Section~\ref{sec:alg} ($\blackbox$ is constructed similarly), and how we perform the user studies in Section~\ref{sec:exp}.

\section{Theoretical Framework}
\label{sec:theory}

We define notions of a potentially untrustworthy black box $\blackbox$ and a potentially misleading explanation $\explanation$ for $\blackbox$. These notions are only used to guide our algorithms; once we have constructed $\blackbox$ and $\explanation$, we test whether $\blackbox$ is actually untrustworthy and $\explanation$ is actually misleading via user studies. Finally, we discuss when potentially misleading explanations exist.

\para{Quantifying user trust.}
We consider a simple approach to estimating whether a user trusts $\blackbox$ given $E$. We assume their key criterion is which features are included in $E$ and which ones are omitted. More precisely, we assume the feature space can be decomposed into $\X=\XD\times\XA\times\XP$, where $\XD$ corresponds to the \textbf{\emph{desired features}} $D$ that the user expects to be included, $\XA$ corresponds to the \textbf{\emph{ambivalent features}} $A$ for which the user is indifferent about whether they are included, and $\XP$ corresponds to the \textbf{\emph{prohibited features}} $P$ that the user expects to be omitted.

Next, an \textbf{\emph{acceptable explanation}} $\explanation\in\explanations_+\subseteq\explanations$ is one where desired features appear in $\explanation$ and the prohibited features do not. Then, we estimate that user decisions $\OO(\explanation)$ are based on (i) whether $\explanation$ is acceptable, and (ii) whether $\explanation$ meets a minimum level $\epsilon_+\in\mathbb{R}_{\ge0}$ of fidelity---i.e., defining
\begin{align*}
\hat\OO(\explanation)=\mathbb{I}[\explanation\in\explanations_+\wedge L(\explanation,\blackbox)\le\epsilon_+],
\end{align*}
we have estimate $\hat\OO(\explanation)\approx\OO(\explanation)$. Similarly, for black boxes that are interpretable, an \textbf{\emph{acceptable blackbox}} $\blackbox\in\blackboxes_+\subseteq\blackboxes$ is one where the desired features appear in $\blackbox$ and the prohibited features do not. Then, we estimate that user decisions $\OO^*(\blackbox)$ are based on whether $\blackbox$ is acceptable---i.e., letting $\hat\OO^*(\blackbox)=\mathbb{I}[\blackbox\in\blackboxes_+]$, we have $\hat\OO^*(\blackbox)\approx\OO^*(\blackbox)$. The user studies we perform demonstrate that $\hat\OO$ and $\hat\OO^*$ are good estimates of $\OO$ and $\OO^*$, respectively; see Section~\ref{sec:exp}.

Now, we say $\blackbox$ is \textbf{\emph{potentially untrustworthy}} if $\hat\OO^*(\blackbox)=0$, and say $E$ is \textbf{\emph{potentially misleading}} if $\hat\OO(\explanation)\neq\hat\OO^*(\blackbox)$. Figure~\ref{fig:blackboxandexp} shows a potentially untrustworthy blackbox (left) and a potentially misleading explanation (right).

%

\para{Existence of potentially misleading explanations.}
We study when potentially misleading explanations exist. 
First, even if an explanation has perfect fidelity, it can still be potentially misleading:
\begin{theorem}
\label{thm:exist}
There exists a black box $\blackbox$ and an explanation $\explanation$ of $\blackbox$ such that (i) $\explanation$ has perfect fidelity (i.e., $L(\explanation,\blackbox)=0$), and (ii) $\explanation$ is potentially misleading. [See Appendix~\ref{sec:apptheory} for proof]
\end{theorem}
\hide{
\vspace{-0.12in}
\begin{proof}
See Appendix~\ref{sec:apptheory}
\end{proof}
}
This result is for a specific black box and a specific explanation of that black box. Next, we study more general settings where potentially misleading explanations exist. Let $\explanation\in\explanations$ be the best explanation for black box $\blackbox$. We focus on the case where $\hat\OO^*(\blackbox)=0$ (i.e., the black box is potentially untrustworthy), so $\explanation$ is potentially misleading if $\hat\OO(\explanation)=1$. Intuitively, potentially misleading explanations exist when the prohibited features $P$ can be \emph{reconstructed} from the remaining ones $D\cup A$. In this case, a misleading explanation can internally reconstruct $P$ using the $D\cup A$. A potential concern is that even when $P$ can be reconstructed, it may not be possible to do so using an interpretable model. We show that an acceptable interpretable model can reconstruct $P$ as long as (i) an acceptable black box $\blackbox_+$ can reconstruct $P$ and achieve good accuracy, and (ii) we can explain $\blackbox_+$ using an acceptable interpretable model that achieves high fidelity. Intuitively, we expect (i) to hold when $P$ can be reconstructed from $D\cup A$, and we expect (ii) to hold since an explanation of $\blackbox_+$ should not depend on features not in $\blackbox_+$.

We formalize (i) and (ii). For (i),
let $\blackbox_+\in\blackboxes_+$ be the best acceptable blackbox. The \emph{restriction error} is
$\epsilon_R=L(\blackbox_+,\blackbox)$.
Then, (i) corresponds to $\epsilon_R\approx0$---i.e., $P$ can be reconstructed from $D\cup A$ when $\blackbox_+$ can then achieve loss similar to $\blackbox$ by internally reconstructing $P$. For (ii), let $\explanation'\in\explanations$ be the best explanation for $\blackbox_+$, and let $\explanation_+\in\explanations_+$ be the best acceptable explanation of $\blackbox_+$. The \emph{acceptable relative error} is the gap in fidelity between these two---i.e.,
\begin{align*}
\epsilon_A=L(\explanation',\blackbox_+)-L(\explanation_+,\blackbox_+)\ge0.
\end{align*}
Then, (ii) corresponds to $\epsilon_A\approx0$---i.e., $\explanation_+$ is almost as good an explanation of $\blackbox_+$ as $\explanation'$. Intuitively, this assumption should hold since $\blackbox_+$ does not use $P$, so there should exist a high fidelity explanation of $\blackbox_+$ that does not use $P$.

Finally, suppose that $\epsilon_R,\epsilon_A$ are small, and that there exists a high fidelity explanation $\explanation\in\explanations$ (which may not be acceptable); then, $\explanation_+$ is potentially misleading:
\begin{theorem}
\label{thm:main}
Suppose $\OO^*(B)=0$; if
$L(\explanation,\blackbox)+2\epsilon_R+\epsilon_A\le\epsilon_+$,
then $\explanation_+$ is potentially misleading. [See Appendix~\ref{sec:apptheory} for proof]
\end{theorem}
\hide{
\vspace{-0.25in}
\begin{proof}
See Appendix~\ref{sec:apptheory}
\end{proof}
}
\section{Generating Misleading Explanations}
\label{sec:alg}


Our algorithm for constructing misleading explanations of black boxes builds on the Model Understanding through Subspace Explanations (MUSE) framework~\cite{lakkaraju19faithful} 
by incorporating additional constraints that enable us to output high fidelity explanations that include desired features and omit prohibited features.

\subsection{Background on MUSE}

Given a black box, MUSE produces an explanation in the form of a \emph{two-level decision set}, which intuitively is a model consisting of nested if-then statements where the nesting depth is two. MUSE chooses an explanation that maximizes two objectives: (i) interpretability: easier for humans to understand, and (ii) fidelity: the explanation should mimic the behavior of the black box.

\para{Two-level decision sets.}
A two-level decision set $R:\mathcal{X}\to\mathcal{Y}$ is a hierarchical model consisting of a set of decision sets, each of which is embedded within an outer if-then structure.
\footnote{The clauses within each of the two levels are unordered, so multiple rules may apply to a given example $x\in\mathcal{X}$. Ties between different if-then clauses are broken according to which rules are most accurate; see~\cite{lakkaraju19faithful} for details.}
Intuitively, the outer if-then rules can be thought of as \emph{neighborhood descriptors} which correspond to different parts of the feature space, and the inner if-then rules are patterns of model behaviors within the corresponding neighborhood. Formally, a two-level decision set has form
\begin{align*}
R = \{ (q_1, s_1, c_1), \cdots, (q_M, s_M, c_M)\},
\end{align*}
where $c_i\in\mathcal{Y}$ is a label, and $q_i$ and $s_i$ are conjunctions of predicates of the form ``$\text{feature}\sim\text{value}$'', where $\sim\;\in\{=,\ge,\le\}$ is an operator; e.g., ``$\text{age}\geq50$'' is a predicate. In particular, $q_i$ corresponds to the neighborhood descriptor, and $(s_i,c_i)$ together represent the inner if-then rules with $s_i$ denoting the antecedent (i.e., the if condition) and $c_i$ denoting the consequent (i.e., the corresponding label). 


\para{Optimization problem.}
Below, we give an overview of the objective function of MUSE. The objective of MUSE is estimated on a given training dataset $\mathcal{D}$ in the context of a two-level decision set $R$ and a black box $\blackbox$.

First, there are many measures of interpretability---e.g., explanations with fewer rules are typically easier to understand. MUSE employs seven such measures. The first four measures are the number of predicates $f_1(R)$, the feature overlap $f_2(R)$,
the rule overlap $f_3(R)$,
and the cover $f_4(R)$;
these four measures are part of the optimization objective. The next three measures are the size $g_1(R)$,
the maximum width $g_2(R)$,
and the number of unique neighborhood descriptors $g_3(R)$; these three measures are included as constraints in the optimization problem. For details on the definitions of these measures, see Appendix~\ref{sec:appinterp}.

Second, fidelity is measured as before---e.g., the accuracy relative to $\blackbox$. We use $f_5(R)$ to denote the fidelity of $R$.

Finally, to construct the search space, we use frequent itemset mining (e.g., apriori~\cite{agrawal1994fast}) to generate two sets of potential if conditions (i.e., sets of conjunctions of predicates): (i) $\mathcal{ND}$ from which we can choose the neighborhood descriptors, and (ii) $\mathcal{DL}$ from which we can choose the inner if-then rules. Then, the complete optimization problem is:
\begin{align}
\label{eq:opt}
& \argmax_{R \subseteq \mathcal{ND} \times \mathcal{DL} \times \mathcal{C}} ~ \sum\limits_{i=1}^{5} \lambda_i f_i(R) \\
& \text{subj. to } g_i(R) \le \epsilon_i ~ \forall i\in\{1,2,3\}. \nonumber
\end{align}
The hyperparameters $\lambda_1,...,\lambda_5\in\mathbb{R}_{\ge0}$ can be chosen using cross-validation; $\epsilon_1, \epsilon_2, \epsilon_3$ must be chosen by the user.

\para{Optimization procedure.}
The optimization problem (\ref{eq:opt}) is non-normal, non-negative, non-monotone, and submodular with matroid constraints~\cite{lakkaraju19faithful}. Exactly solving this problem is NP-Hard~\cite{khuller1999budgeted}. Approximate local search provides the best known theoretical guarantees for this class of problems---i.e., $(k+2+1/k+\delta)^{-1}$, where $k$ is the number of constraints and $\delta > 0$~\cite{lee2009non}. 

\subsection{Our Approach}
We extend MUSE to generate potentially misleading explanations by modifying the optimization problem (\ref{eq:opt}). In particular, we need to (i) ensure that none of the prohibited features $P$ (e.g., race) appear in the explanation (even if they are being used by the black box to make predictions), and (ii) ensure that all the desired features $D$ appear (even if they are not being used by the black box).
Formally, let $\mathcal{ND}_+\subseteq\mathcal{ND}$ denote the set of candidate if conditions for outer if clauses that do not include any prohibited attributes, and let $\mathcal{DL}_+\subseteq\mathcal{DL}$ be the analog for inner if clauses.
Furthermore, we also add a term to the objective that measures the number of features in $\XD$ that are part of some rule in $R$:
\begin{align*}
\text{coverdesired}(R) = \sum_{d \in D} \mathbb{I}[\exists (q, s, c) \in R \text{ s.t. } d \in (q \cup s)],
\end{align*}
where $d\in D$ is a desired feature. Maximizing this value will in turn maximize the chance that every desired attribute appears somewhere in the explanation.

Together, we use the following optimization problem to construct candidate misleading explanations:
\begin{align}
\label{eq:optnew}
& \argmax_{R \subseteq \mathcal{ND}_+ \times \mathcal{DL}_+ \times \mathcal{C}} ~ \sum\limits_{i=1}^{5} \lambda_i f_i(R) + \lambda_6f_6(R) \\
& \text{subj. to } g_i(R) \le \epsilon_i ~ (\forall i\in\{1,2,3\}) \nonumber
\end{align}
where $f_6(R)=\text{coverdesired}(R)$. The following theorem shows that as before, we can solve (\ref{eq:optnew}) with approximate local search:
\begin{theorem}
\label{thm:submodular}
(\ref{eq:optnew}) is non-normal, non-negative, non-monotone, and submodular, and has matroid constraints. [See Appendix~\ref{sec:apptheory} for proof]
\end{theorem}
\hide{
\vspace{-0.15in}
\begin{proof}
See Appendix~\ref{sec:apptheory}
\end{proof}
}
\section{Experimental Evaluation}
\label{sec:exp}

Our goal is to evaluate how explanations can affect users' trust of a black box. To this end, we first construct a black box and its explanations. Then, we perform a user study with domain experts to understand how each explanation affects user trust of the black box. All of our experiments are performed in the context of a real world application - bail decisions.

A key aspect of our approach is that the ``black box'' $\blackbox$ that we construct is itself an interpretable model.
This allows us to evaluate whether $\blackbox$ is actually untrustworthy (i.e., $\OO^*(\blackbox)=0$) via user studies.
\footnote{For the user study checking $\OO(\explanation)$, we do not show users the internals of $\blackbox$, so their decision of whether to trust $\blackbox$ is not affected by the fact that $\blackbox$ happens to be interpretable.}
Also, for an explanation $\explanation$ of $\blackbox$, we can check if $\blackbox$ is trusted given only on $\explanation$ (i.e., $\OO(\explanation)=1$). If both of these criteria hold i.e., $\OO^*(\blackbox)\neq\OO(\explanation)$, then explanation $\explanation$ is misleading.


\para{Bail decisions.}
Our experiments focus on bail decision making, a high-stakes task. Police arrest over 10 million people each year in the U.S.~\cite{kleinberg2017human}. Soon after arrest, judges decide whether defendants should be released on bail or must wait in jail until their trial. Since cases can take several months to proceed to trial, bail decisions are consequential both for defendants as well as society. By law, 
a defendant should be released only if the judge believes that they will not flee or commit another crime. This decision is naturally modeled as a prediction problem.

We use a dataset on bail outcomes collected from several state courts in the U.S. between 1990-2009~\cite{lakkaraju19faithful}. This dataset contains 37 features, including demographic attributes (age, gender, race), personal (e.g., married) and socio-economic information (e.g., pays rent, lives with children), current offense details (e.g., is felony), and past criminal records of about 32K defendants who were released on bail. Each defendant in the data is labeled either as risky (if he/she either fled and/or committed a new crime after being released on bail) or non-risky. The goal is to train a black box that predicts these outcomes to help judges make bail decisions. Explanations of this black box
are needed to help domain experts determine whether to trust the black box.

\para{Domain experts in user study.}
We carried out our study with 47 subjects. Each participant is a student enrolled in a law school
at the time of our study. Each participant acknowledged having in-depth knowledge (16 participants) or at least some familiarity (31 participants) with the bail decision making process. Of the subjects, 27 self-identified as male and 20 as females; 25 are White, 15 Asian, 2 Hispanic, and 5 African American. 

We split our study into two phases: (i) First, we reached out to each of the participants to determine which of the features in the bail dataset are relevant (i.e., desired) and which ones should be omitted (i.e., prohibited). We used these insights to construct our classifier and its explanations (see Section~\ref{sec:expalg}). (ii) Next, we performed the key part of our study---we reached out to all the subjects to understand how/why a particular explanation influences their trust of the black box classifier.

\begin{figure}
\centering
\includegraphics[width=0.23\textwidth]{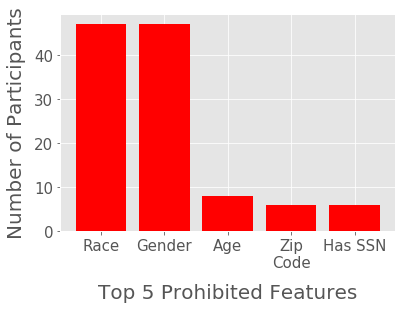}
\includegraphics[width=0.23\textwidth]{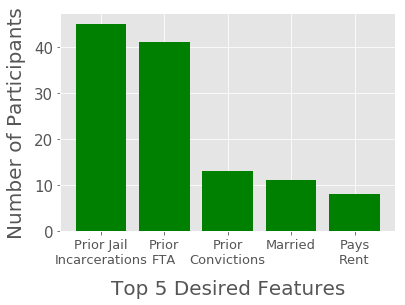}
\caption{Top 5 prohibited (left) and desired features (right), and number of participants who voted for each one.}
\label{fig:survey}
\vspace{-0.2in}
\end{figure}

\subsection{Constructing the Black Box and Explanations}
\label{sec:expalg}
We discuss how we construct our black box (designed to be untrustworthy) and its explanations (some of which are designed to be misleading). We surveyed the domain experts to identify desired and prohibited features, and then used this information to construct our classifier and explanations. We generate an untrustworthy black box $\blackbox$ by explicitly including prohibited features and omitting desired features, and generate misleading explanations for $\blackbox$ by explicitly including desired features and/or omitting prohibited features. 

\para{Identifying prohibited and desired features.}
We surveyed all our 47 subjects to identify prohibited and desired features. Each participant is shown all 37 features in the bail dataset, and is asked to indicate which ones are relevant and which ones should be omitted when predicting if a defendant is risky and should not be released on bail. Figure~\ref{fig:survey} shows the 5 features ($x$-axis) ranked as the most prohibited (left) and the most desired (right) ones by the participants. It also shows how many participants voted for each feature ($y$-axis). Race and gender stand out unanimously as the top prohibited features; prior jail incarcerations (PJI) and prior failure to appear (PFTA)
\footnote{If a defendant has failed to appear in the past, that means they failed to show up for court dates and is deemed a flight risk.}
are the top desired features. In both cases, the first two features received significantly more votes compared to all the other features, so we use race and gender as prohibited features, and use PJI and PFTA as desired features in all subsequent experiments.

\para{Black box and explanations.}
%
We use the identified prohibited and desired features to construct our black box and its explanations. At a high level, our approach is to construct a black box that is designed to be untrustworthy to the domain experts should they be familiar with its inner workings, and construct high-fidelity explanations of this black box designed to mislead them into trusting the black box.

To this end, we randomly shuffle the bail dataset and split it into train (70\%), test (25\%), and validation (5\%) sets. We employ our framework with different parameter settings to construct both the black box and its explanations. We leverage the validation set and a coordinate descent style tuning procedure similar to that of MUSE to set the hyperparameters $\lambda_1, \lambda_2, ..., \lambda_6$~\cite{lakkaraju19faithful}.

We first construct a black box $\blackbox$ that uses race and gender (prohibited) and does not use PJI and PFTA (desired); thus, $\blackbox$ is most likely untrustworthy to the domain experts should they examine its internal workings. We use our framework to build $\blackbox$; while designed to construct explanations, it can be applied to build an interpretable classifier by replacing the black box labels $\blackbox(x)$ (for each $x\in\mathcal{X}$) with the corresponding ground truth label $y$.
We use desired features $D = \{\text{PJI},\text{PFTA}\}$ and prohibited features $P = \{\text{race},\text{gender}\}$.
The resulting black box $\blackbox$, shown in  Figure~\ref{fig:blackboxandexp} (left), is an interpretable two-level decision set; its accuracy on the held-out test set is 83.28\%. 

We then use our framework to construct three different high-fidelity explanations $\explanation_1, \explanation_2, \explanation_3$ of $\blackbox$, as follows: (i) $\explanation_1$ does not use either prohibited features or desired features (i.e., we use $P = \{\text{race}, \text{gender}, \text{PJI}, \text{PFTA}\}$ and $D = \varnothing$), (ii) $E_2$ uses both prohibited and desired features (i.e., we use $P = \varnothing$ and $D = \{\text{race}, \text{gender}, \text{PJI}, \text{PFTA}\}$, and (iii) $E_3$ uses desired features but not prohibited features (i.e., we use $P = \{\text{race}, \text{gender}\}$ and $D = \{\text{PJI}, \text{PFTA}\}$. We show $\explanation_3$ in Figure~\ref{fig:blackboxandexp} (right); 

A potential concern is that our goal is to study how qualititative aspects of each explanation (e.g., which features appear) affects whether a user trusts $\blackbox$; however, the fidelity of an explanation can also affect user trust. Thus, it is important to control for fidelity beforehand.
To this end, we estimate the fidelity of each explanation on the held-out test set; the fidelities for $E_1, E_2, E_3$ are 97.3\%, 98.9\%, and 98.2\% respectively. These values are all very similar; thus, differences in whether the user trusts or mistrusts $\blackbox$ must be due to the structure of the explanations rather than their fidelities.


\begin{figure}[!t]
\center{\includegraphics[width=0.40\textwidth]{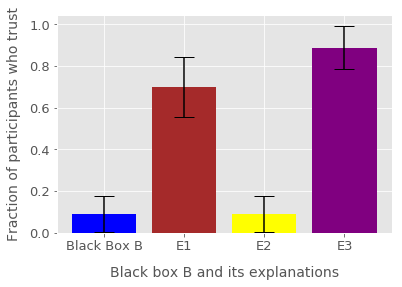}}
\vspace{-0.1in}
\caption{\label{fig:mainresult} Effect of various explanations on user trust of black box $\blackbox$.}
\vspace{-0.2in}
\end{figure}

\subsection{Human Evaluation of Trust in Black Box}
\label{sec:evaluation}
Next, we performed a user study with the domain experts to understand how our different explanations $E_1, E_2, E_3$ affect user trust of the same black box model $\blackbox$. 

\para{User study design.}
We designed an online user study in which 41 of the 47 domain experts that we recruited participated.
\footnote{Remaining 6 participants were used to explore how interactive explanations can affect user trust.}
Each participant was randomly chosen to be shown either the black box $\blackbox$ (with fidelity 100\%) or one of the explanations $\explanation_1,\explanation_2,\explanation_3$ (with their corresponding fidelities).
Including the black box $\blackbox$ is critical since it allows us to estimate the baseline trust $\OO^*(\blackbox)$---i.e., whether users trust $\blackbox$ if they understand its internals. Each participant was instructed beforehand that the explanations they see are only correlational, not causal. Participants were allowed to take as much time as they wanted to complete the study. 

Each participant was asked (i) to answer the following yes/no question: \emph{``Below is an explanation generated by state-of-the-art ML for a particular black box designed to assist judges in bail decisions. Based on this explanation, would you trust the underlying model enough to deploy it?"}, and (ii) a follow-up descriptive question to explain \emph{why} they decided to trust or mistrust the black box. 

\para{Results and discussion.}
Figure~\ref{fig:mainresult} shows the results of our user study. Each of the bars corresponds to either the black box or one of the explanations ($x$-axis). We show the corresponding user trust, measured as the fraction of participants who responded that they trust the underlying black box---i.e., answered yes to the question above ($y$-axis). 

As can be seen, only 9.1\% of the participants who saw the actual black box trusted it (blue), establishing our baseline that the black box is not trustworthy. Next, we discuss users who only saw one of the explanations of the black box. First, only 10\% of the participants who saw $E_2$ (brown), which includes race and gender as well as PJI and PFTA, trusted the underlying black box. On the other hand, 70\% and 88\% of participants who saw $E_1$ (yellow) and $E_3$ (purple), respectively, trusted the underlying black box. The prohibited features race and gender do not appear in $E_1$ or $E_3$; in addition, $E_3$ includes the desired features PJI and PFTA.

These results show that $E_1$ and $E_3$ are misleading users---i.e., they lead the user to trust a black box, while users find the actual black box untrustworthy. Since $\blackbox$ and $E_2$ both include race and gender, participants are unwilling to trust the black box in these two cases. On the other hand, race and gender do not appear in $E_1$ and $E_3$, and in these cases users are very likely to trust the underlying black box. These results are in spite of the clear warning we show to participants saying that the explanations shown are not causal. Furthermore, participants who see $E_3$ appear to trust the underlying black box more frequently than those who see $E_1$, most likely since the desired attributes PJI and PFTA are used by $E_3$.

Finally, we analyzed the reasons participants gave for their responses. They are consistent with our findings---i.e., user trust appears to primarily be driven by whether the race and gender features appear in the explanation shown. 
\vspace{-0.1in}
\section{Discussion \& Conclusions}

We carried out the first systematic study of if and how explanations of black boxes can mislead users and affect user trust, including a novel theoretical framework for understanding when misleading explanations can exist,
a novel approach for generating explanations that are likely to be misleading,
and an extensive user study with domain experts from law and criminal justice to understand how misleading explanations impact user trust. We find that user trust can be manipulated by high-fidelity, misleading explanations.
These misleading explanations exist since prohibited features (e.g., race or gender) can be reconstructed based on correlated features (e.g., zip code). Thus, adversarial actors can fool end users into trusting an untrustworthy black box---e.g., one that employs prohibited attributes to make decisions.
We consider two ways to address this challenge.

%
First, recent research~\cite{lakkaraju19faithful} has advocated for thinking about explanations as an interactive dialogue where end users can query or explore different explanations (called \emph{perspectives}) of the black box. In fact, MUSE is designed for interactivity---e.g., a judge can ask MUSE ``How does the black box make predictions for defendants of different races and/or genders?", and it would return an explanation that only uses race and/or gender on outer if-then clauses. We performed another user study with 6 domain experts from our participant pool to study their trust in the underlying black box $\blackbox$ when they could explore various explanations of $\blackbox$ using MUSE, and found that only 16.7\% of the participants (1 out of 6) trusted $\blackbox$. This value is much closer to the baseline trust (9.1\%). 

%
Second, there has been recent work on capturing causal relationships between input features and black box predictions~\cite{zhao2019causal,wachter2017counterfactual}. Explanations relying on correlations not only may be misleading~\cite{rudin2019stop}, but have also been shown to lack robustness~\cite{ghorbani2019interpretation}, and causal explanations may address these issues. 

\bibliographystyle{aaai}
\bibliography{paper}

\clearpage
\appendix
\section{Algorithm}

\subsection{Interpretability Measures}
\label{sec:appinterp}

We describe how each of our interpretability objectives are measured given a two-level decision set $R$ (with $M$ rules), a black box $\blackbox$, and a training set $\mathcal{D} = \{x_1, ..., x_N\} \subseteq \mathcal{X}$. These measures are summarized in Table~\ref{quantify}.

The first four measures are structural properties of $R$. First, we want to minimize the size of $R$, which is the number of triples $(q,s,c)$ in $R$. Second, we want to minimize the \emph{maximum width} of $R$, which is the maximum over $(q,s,c)$ in $R$ of the quantities $\text{width}(s)$ and $\text{width}(q)$, where the width of an if-then rule is the number of predicates that occur in its condition. Third, we want to minimize the total number of predicates in $R$, which is the sum over $(q,s,c)$ in $R$ of $\text{width}(s)+\text{width}(q)$.
Fourth, we want to minimize the number of decision sets in $R$, which is the number of unique neighborhood descriptors $q$ in $\mathcal{R}$.

The next measures are are semantic properties of $R$. Intuitively, this objective captures the idea that outer if-then clauses (i.e., neighborhood descriptors) and inner if-then rules have different semantic meanings. To make the distinction more clear, the overlap between the features that appear in outer and inner if-then rules should be minimized. 
In particular, for each pair $(q,s)$ of outer and inner if-then rules, we sum up the number of features that occur in both $q$ and $s$; we want to minimize this quantity.

The final measure captures a property specific to decision sets. For decision sets, multiple rules may apply for a given example $x\in\mathcal{X}$;
\footnote{A rule applies to $x$ if $x$ satisfies its condition.}
i.e., rules can be \emph{ambiguous}. To maximize interpretability, for most examples $x$, the rules should be unambiguous---i.e., only one rule should apply for a given $x$. First, we want to minimize the \emph{rule overlap}, which is the number of extra rules that apply. Second, we want to maximize \emph{cover}, which counts the number of instances in the dataset that satisfy some rule in $\mathcal{R}$.

Finally, we have
\begin{align*}
f_1(\mathcal{R}) & = 2 \mathcal{W}_{\text{max}}\cdot |\mathcal{ND}| \cdot |\mathcal{DL}| - \text{numpreds}(R) \\
f_2(\mathcal{R}) &= \mathcal{W}_{\text{max}} \cdot |\mathcal{ND}| \cdot |\mathcal{DL}| - \text{featureoverlap}(R) \\
f_3(\mathcal{R}) &= N (|\mathcal{ND}| \cdot |\mathcal{DL}|)^2 - \text{ruleoverlap}(R)  \\ 
f_4(\mathcal{R}) &= \text{cover}(R) \\
f_5(\mathcal{R}) &= N \cdot |\mathcal{ND}| \cdot |\mathcal{DL}|  - \text{disagreement}(R)
\end{align*}
\normalsize
where ${W}_{\text{max}}$ is the maximum width of any rule in either candidate sets. 
To ensure that the objective is non-negative, we have subtracted each measure from its upper bound.

\section{Proofs of Theorems}
\label{sec:apptheory}

\para{Proof of Theorem~\ref{thm:exist}.}
Consider input features $\XD=\XP=\mathbb{R}$, and there are no ambivalent features, so $\mathcal{X}=\XD\times\XP=\mathbb{R}^2$, and binary labels $\mathcal{Y}=\{0,1\}$. Furthermore, consider a distribution $p((x_1,x_2),y)$ over $\mathcal{X}\times\mathcal{Y}$ defined by
\begin{align*}
p((x_1,x_2),y)=p_0(x_1)\cdot\delta(x_2-x_1)\cdot\delta(y-\mathbb{I}[x_2\ge0]),
\end{align*}
where $p_0=\mathcal{N}(0,1)$. In other words, $x_1$ is a standard Gaussian random variable, $x_1$ and $x_2$ are perfectly correlated, and the outcome is 1 if $x_2\ge0$ and 0 otherwise. Next, consider a black box
\begin{align*}
\blackbox((x_1,x_2))=\mathbb{I}[x_2\ge0],
\end{align*}
i.e., $\blackbox$ achieves zero loss. Since $\blackbox$ uses the prohibited feature $x_2$, it is probably untrustworthy---i.e., $\hat\OO^*(\blackbox)=0$. Similarly, consider an explanation
\begin{align*}
\explanation((x_1,x_2))=\mathbb{I}[x_1\ge0].
\end{align*}
Since this explanation uses the desired feature and not the prohibited feature, it is acceptable; thus, it is probably misleading---i.e., $\hat\OO(\explanation)\neq\hat\OO^*(\blackbox)$. Finally, note that
\begin{align*}
&L(\explanation,\blackbox) \\
&=\mathbb{E}_{p((x_1,x_2))}[\ell(\explanation((x_1,x_2)),\blackbox((x_1,x_2))] \\
&=\mathbb{E}_{p((x_1,x_2))}[\ell(\mathbb{I}[x_1\ge0],\mathbb{I}[x_2\ge0])] \\
&=\mathbb{E}_{p(x_1)}\left[\int\ell(\mathbb{I}[x_1\ge0],\mathbb{I}[x_2\ge0])\cdot\delta(x_2-x_1)dx_2\right] \\
&=\mathbb{E}_{p(x_1)}[\int\ell(\mathbb{I}[x_2\ge0],\mathbb{I}[x_2\ge0])] \\
&=0.
\end{align*}
Thus, $\explanation$ achieves perfect fidelity, as claimed. $\qed$

\begin{table}[t]
\caption{Summary of notation.}
\scriptsize
\centering
\begin{tabular}{|c|l|} 
 \hline
Symbol & Description \\
\hline
$\mathcal{D}$ & Dataset $\mathcal{D} = \{x_1, x_2, ..., x_N\}$\\
$\mathcal{ND}$ & Candidate set of conjunctions of predictions for choosing \\ & outer if-then clauses of explanations\\
$\mathcal{DL}$ & Candidate set of conjunctions of predictions for choosing \\ & inner if-then rules of explanations \\
\hline
\end{tabular}
\end{table}

\begin{table}
\caption{Metrics used in the optimization problem.}
\tiny
\begin{tabular}{| l | l | }
\hline
fidelity & $\text{disagreement}(R) = \sum\limits_{i=1}^{M} | \{ x \in \mathcal{D} \mid x \text{ satisfies } q_i \wedge s_i, ~ \blackbox(x) \neq c_i \} |$ \\
\hline
\multirow{12}{*}{interpretability} & $\text{size}(R) =\text{ number of triples } (q,s,c) \text{ in }R$ \\
& $\text{maxwidth}(R) = \max\limits_{ e \in \bigcup\limits_{i=1}^{M} (q_i \cup s_i)}\text{width}(e)$ \\
& $\text{numpreds}(R) = \sum\limits_{i=1}^{M} \text{width}(s_i) + \text{width}(q_i)$ \\
& $\text{numdsets}(R) = |\text{dset}(R)| \text{ where } \text{dset}(R) = \bigcup\limits_{i=1}^{M} q_i$ \\
& $\text{featureoverlap}(R) = \sum\limits_{q \in \text{dset}(R)} \sum\limits_{i=1}^{M} \text{featureoverlap}(q, s_i)$ \\
\hline
\multirow{3}{*}{unambiguity} & $\text{ruleoverlap}(R) = \sum\limits_{i=1}^{M} \sum\limits_{j=1, j \neq i}^{M}\text{overlap}(q_i \wedge s_i, q_j \wedge s_j)$ \\
 & $\text{cover}(R) = | \{ x \in \mathcal{D} \mid x \text{ satisfies } q_i \wedge s_i \text{ for some } i \in \{1 \cdots M\} \} | $ \\
\hline
\end{tabular}
\normalsize
\label{quantify}
\end{table}

\para{Proof of Theorem~\ref{thm:main}.}
First, we have the following decomposition of the relative error: for any $F,F',F'':\mathcal{X}\to\mathcal{Y}$,
\begin{align*}
L(F,F')\le L(F,F'')+L(F'',F').
\end{align*}
This result follows since for any $y,y',y''\in\mathcal{Y}$,
\begin{align*}
\ell(y,y')
&=\mathbb{I}[y\neq y'] \\
&=\mathbb{I}[y\neq y''\wedge y''=y']+\mathbb{I}[y=y''\wedge y''\neq y'] \\
&\le\mathbb{I}[y\neq y'']+\mathbb{I}[y''\neq y'] \\
&=\ell(y,y'')+\ell(y'',y'),
\end{align*}
so we have
\begin{align*}
L(F,F')
&=\mathbb{E}_{p(x)}[\ell(F(x),F'(x))] \\
&\le\mathbb{E}_{p(x)}[\ell(F(x),F''(x))+\ell(F''(x),F'(x))] \\
&=L(F,F'')+L(F'',F').
\end{align*}
As a consequence, we have
\begin{align*}
L(\explanation,\blackbox)
&\le L(\explanation,\blackbox_+)+L(\blackbox_+,\blackbox) \\
&= L(\explanation,\blackbox_+)+\epsilon_R.
\end{align*}
Next, note that
\begin{align*}
L(\explanation,\blackbox_+)
&\le L(\explanation',\blackbox_+)+L(\blackbox_+,\blackbox) \\
&=L(\explanation_+,\blackbox_+)+L(\blackbox_+,\blackbox)+\epsilon_A,
\end{align*}
where the first line follows since by definition, $\explanation'$ maximizes error relative to $\blackbox_+$ over $\explanation\in\explanations$, and the second line follows by the definition of $\epsilon_A$. Now, again by our decomposition of relative error, we have
\begin{align*}
L(\explanation_+,\blackbox_+)
&\le L(\explanation_+,\blackbox)+L(\blackbox,\blackbox_+) \\
&=L(\explanation_+,\blackbox)+\epsilon_R,
\end{align*}
where the last line follows since relative error is symmetric. Putting these three inequalities together, we have
\begin{align*}
L(\explanation_+,\blackbox)
&\le L(\explanation,\blackbox)+2\epsilon_R+\epsilon_A \\
&\le\epsilon_+,
\end{align*}
where the second line follows by our assumption in the theorem statement. Since $\explanation_+\in\explanations_+$, by definition of $\hat\OO$, we have $\hat\OO(\explanation_+)=1$, as claimed. $\qed$.

\para{Proof of Theorem~\ref{thm:submodular}.}
If at least one term in a linear combination is non-normal (resp., non-monotone), then the entire linear combination is non-normal (resp., non-monotone). Given that the objective in (\ref{eq:opt}) is already non-normal (resp., non-monotone), then it follows that the objective in (\ref{eq:optnew}) it is likewise non-normal (resp., non-monotone). In particular, coverdesired computes how many of the desired features $D$ appear in $R$. By definition, this value cannot be negative. Since the objective in (\ref{eq:opt}) is non-negative and $\text{coverdesired}(R)$ is non-negative, so the objective in (\ref{eq:optnew}) is also non-negative. The non-monotone property follows similarly. Next, we did not add any new constraints to (\ref{eq:optnew}), and the constraints in (\ref{eq:opt}) are known to follow a matroid structure. Thus, (\ref{eq:optnew}) also has matroid constraints.

Finally, note that coverdesired denotes the number of desired features that appear in $R$. This function clearly has diminishing returns---i.e., more desired attributes will be covered when we add a new rule to a smaller set of rules compared to a larger set. Therefore, this function is submodular. Since the objective in (\ref{eq:opt}) is submodular and coverdesired is submodular, it follows that the objective in (\ref{eq:optnew}) is also submodular since a linear combination of submodular functions is submodular. $\qed$

\end{document}